\pdfoutput=1
\documentclass[final,3p,times]{elsarticle}

\usepackage{graphicx}
\usepackage{amssymb}

\usepackage{hyperref}
\usepackage[draft=true,frozencache]{minted}

\usepackage{lineno}
\usepackage{breakurl}

\journal{arXiv}

\begin{document}
\begin{frontmatter}


\title{OpenEDGAR: Open Source Software for SEC EDGAR Research}



\author{Michael J Bommarito II}
\ead{mike@lexpredict.com}
\author{Daniel Martin Katz}
\ead{dan@lexpredict.com}
\author{Eric M Detterman}
\ead{eric@lexpredict.com}
\address{LexPredict, LLC}

\begin{abstract}
OpenEDGAR is an open source Python framework designed to rapidly construct research databases based on the Electronic Data Gathering, Analysis, and Retrieval (EDGAR) system operated by the US Securities and Exchange Commission (SEC).  OpenEDGAR is built on the Django application framework, supports distributed compute across one or more servers, and includes functionality to (i) retrieve and parse index and filing data from EDGAR, (ii) build tables for key metadata like form type and filer, (iii) retrieve, parse, and update CIK to ticker and industry mappings, (iv) extract content and metadata from filing documents, and (v) search filing document contents.  OpenEDGAR is designed for use in both academic research and industrial applications, and is distributed under MIT License at \url{https://github.com/LexPredict/openedgar}.
\end{abstract}


\end{frontmatter}
\section{Introduction}
\label{section:introduction}
Information disclosure through EDGAR has been a requirement for most publicly listed or registered investment companies for the last quarter century; it now contains terabytes of documents and data including press releases, annual corporate filings, executive employment agreements, asset-backed security (ABS) performance, and investment company holdings.  Over the last two decades, researchers around the world and in many disciplines have analyzed this data to ask and answer many important questions (\cite{PREMUROSO20081, asthana2004differential, huddart2007jeopardy, bushee2005economic, rogers2017run}).  However, despite the breadth of research conducted over two decades, it is still difficult for many scholars to carry out or reproduce research based on EDGAR.

OpenEDGAR is an open source \textsc{Python} framework designed to address these problems of access and reproducibility.  Previously, researchers would independently spend time or money to redevelop the same data retrieval and parsing code over and over.  OpenEDGAR allows the community of researchers and developers to share the cost and benefits of this core functionality, increasing access to this research data and lowering the cost of reproducing important research.  We believe that this data, especially when combined with the increasing number of open source resources for natural language processing and machine learning, can unlock answers for many important research questions (\cite{BirdKleinLoper09, scikit-learn, rehurek_lrec, spacy2}).

\subsection{History}
LexPredict first began archiving and indexing data from EDGAR in 2013 in order to develop corpora of legal and regulatory text for natural language and machine learning tasks.  Over time, these scripts developed into a set of backend services behind a data product, the LexPredict Agreement Database, released in 2015.  On December 30, 2016, the SEC decommissioned FTP distribution and enabled its new HTTPS delivery mechanism; much of OpenEDGAR was rewritten and modernized at this time.  LexPredict announced the open sourcing of OpenEDGAR in May 2018.

\subsection{License and Support}
OpenEDGAR is released under the MIT license, allowing for permissive use commercially and GPL compatibility if desired.  Support is provided through GitHub issue tracking at \url{https://github.com/LexPredict/openedgar/issues}.

\section{Design}
\label{section:design}
OpenEDGAR is designed to provide an open source Python framework for working with EDGAR data at any scale.  This goal is accomplished by building on top of high-quality open source packages and through careful architecture choices that enable researchers to attack problems both large and small.  As we have outlined in allied work (\cite{lexnlp2018}), our guiding package selection and design principles are stated below:

\begin{enumerate}
\item \textbf{Standard open source licensing}: We strongly prefer dependencies with standard open source licensing options like MIT, Apache, or GPL-family licenses.
\item \textbf{High level of maturity}: We strongly prefer dependencies with mature code bases, including years of development and testing.
\item \textbf{High level of documentation}: We strongly prefer dependencies with well-documented code bases.
\item \textbf{Broad language and character support}: We strongly prefer dependencies that natively support non-English as well as English text.
\item \textbf{Strong ecosystem}: We strongly prefer dependencies with large and active communities of developers and users.
\item \textbf{Simple scalability}: We strongly prefer dependencies that support parallel or distributed patterns without complex infrastructure setup.
\end{enumerate}

\subsection{Architecture}
OpenEDGAR is based on standard multi-tier architecture using the \textsc{Django} application framework.  Discussion of each of the tiers of the architecture is provided below.

\begin{enumerate}
\item \textbf{Object Storage}: Researchers who need to search EDGAR must retrieve and store terabytes of data, and new documents continue to be filed every day.  If researchers desire to distribute or parallelize their analysis, then this data must be managed to allow for such access patterns.  Based on these requirements, we have designed OpenEDGAR to use Amazon Simple Storage Service (S3) or compatible object storage engines such as OpenStack Swift.  Raw filings are stored with object keys that match the SEC's own naming scheme.  Filing documents are stored and deduplicated through SHA1-based keying.

\item \textbf{Relational Database}: While raw filings and document contents represent the largest storage requirement for EDGAR, there are a large number of index or metadata records that are important for analysis.  OpenEDGAR uses traditional relational database technology to manage these records, allowing users to interact with this data either through the Django ORM or through SQL directly.  By default, OpenEDGAR uses \textsc{Postgres} as we prefer its mature code base, community, feature set, and license; however, Django also supports many databases such as MySQL, Oracle, or SQLite.

\item \textbf{Distributed Task and Message Queues}: Many researchers find it necessary to distribute work across one or more servers to handle compute or memory requirements.  To meet this requirement, we designed OpenEDGAR to use a distributed and asynchronous task queue, \textsc{celery}, which is most commonly integrated with Django.  For choice of message broker, OpenEDGAR uses RabbitMQ as we prefer its mature code base and feature set; however, celery also supports other brokers.  All key methods in OpenEDGAR are implemented to run either directly or as celery tasks.

\item \textbf{Content Extraction}: EDGAR contains thousands of types of documents, structured and unstructured, across many file formats (\cite{secformindex}).  These documents include normalized XBRL XML filings, procedural narratives like Form 10-K in HTML, PDF, or plain text, open-ended press releases and PowerPoint presentations, and even images containing non-English marketing material.  To handle this heterogeneity, OpenEDGAR relies on the Apache Tika and Tesseract projects to extract and normalize content (\cite{Smith:2007:OTO:1304596.1304846}).  The Apache Tika project is a toolkit for ``detecting and extracting metadata and text from over a thousand different file types,'' and Tika provides a RESTful, parallel Java service.  When Tika encounters images, either embedded in documents or as image PDF files, it can optionally perform optical character recognition (OCR) using Tesseract.  OpenEDGAR does not disable Tesseract support by default, but given the computational expense across the EDGAR corpus, researchers should consider whether this is required for their use case.

\item \textbf{Interactive Data Science Platform}: Some projects eventually develop into traditional web applications.  However, nearly all projects begin through the iterative exploration of data, evolving into more formal analysis or prototypes over time.  OpenEDGAR uses the Jupyter interactive computing platform for this requirement, allowing researchers to interactively develop code in languages like Python or R,  execute in ecosystems like Apache Spark, examine figures and results, and publish source code and results privately or publicly (\cite{Kluyver:2016aa}).
\end{enumerate}

\subsection{Containerization}
Containerization technologies such as Docker are increasingly popular and provide many benefits relative to traditional deployment models.  While most discussion of Docker is focused on scalable deployment of large-scale applications, scholars are increasingly turning to containerization to promote reproducible research (\cite{DBLP:journals/corr/Boettiger14}).  OpenEDGAR will include support and documentation for Docker deployments, as well customization to allow researchers to reproduce specific databases from EDGAR with Dockerfile environment variables.

\subsection{Language Support}
OpenEDGAR is designed to support multiple languages and character sets across its feature set.  While most filings are in English a significant number of filings such as Forms 18-K, 20-K, and 8-K are either written entirely in non-English languages or include portions of non-English text.

\subsection{Unit Testing and Code Coverage}
OpenEDGAR is developed using continuous integration (CI) practices, including unit testing, code coverage analysis, and code style analysis.  Coding style is based on PEP8 and enforced through CI as well.

\section{OpenEDGAR Framework}
\label{section:framework}

OpenEDGAR is built on the \textsc{Django} application framework.  We also use the \textsc{django-cookiecutter} package, as it simplifies the configuration of a number of dependencies and architectural choices.  Django and django-cookiecutter create a large number of boilerplate or template files, which are either required for configuration or necessary for presentation layer code in HTML, CSS, or Javascript.  While these files are distributed in the OpenEDGAR repository, they are not novel to our effort.  Our contributions are detailed below, including the skeleton data model, client code, parsing code, and processing workflows.

\subsection{Data Model}
OpenEDGAR includes a skeleton data model that structures all key metadata provided by the SEC EDGAR system itself.  While most research projects will develop additional requirements and extend the data model, nearly all projects will require these core objects.  A UML diagram of the basic OpenEDGAR data model is provided in Figure \ref{fig:openedgar_skeleton_data_model} and object descriptions are provided below:

\begin{figure}[ht!]
    \centering
    \includegraphics[width=6in]{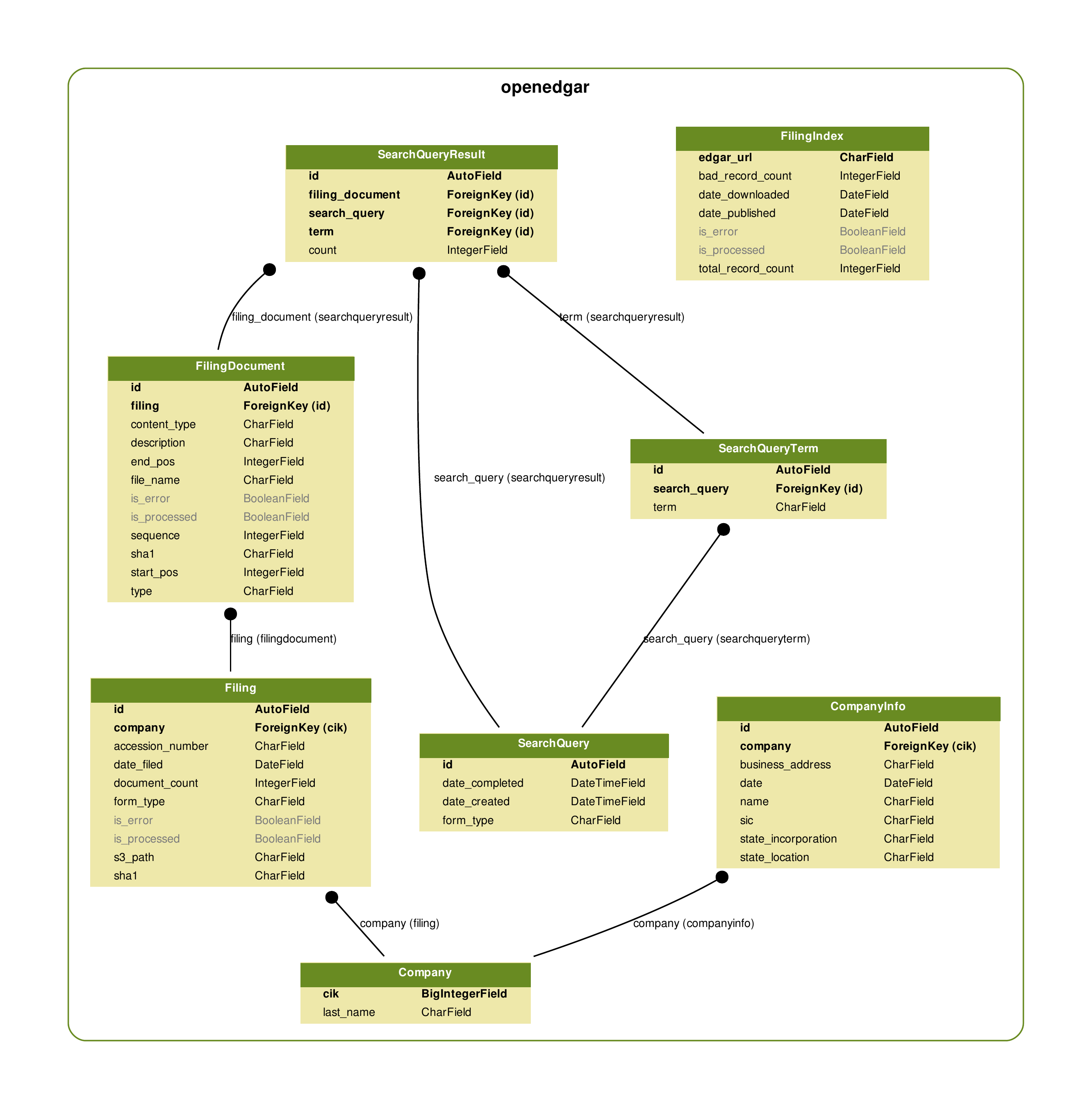}
    \caption{Data model diagram for a skeleton OpenEDGAR search application}
    \label{fig:openedgar_skeleton_data_model}
\end{figure}

\begin{itemize}
    \item \textbf{Company}:  The Company record is the base object for unique Central Index Key values; as such, the name ``Company'' is technically a misnomer, as some CIK values may be assigned to entities or individuals that are not technically companies, such as sovereigns.  However, for ease of use given most use cases, we have retained this name; data model objects can be easily renamed with tools such as PyCharm or sed if desired.
    
    \item \textbf{CompanyInfo}: The CompanyInfo record is the base object for dynamic metadata related to Company records, such as name, state of incorporation, or Standard Industrial Classification (SIC).  This metadata has been solicited from and assigned to filers by the SEC, but often changes over time.
    
    \item \textbf{FilingIndex}: The FilingIndex record is the base object for index files provided by the SEC EDGAR distribution system.  Four different types of indexes are currently provided daily, and ``full'' indexes are compiled for quarterly and annual periods.  Some older index files may be malformed or may not contain all fields, so error handling is required; this table enables efficient handling of state to minimize unnecessary data retrieval and identify sources of errors.
    
    \item \textbf{Filing}: The Filing record is the base object for actual ``filings'' submitted to the SEC.  Filings are keyed by their accession number, which includes both the CIK of the filing agent and an incrementing  identifier.  As with index files, some filings may be malformed or may have been removed the SEC's staff under certain conditions.  This table enables filing search by key metadata such as form type, date, or company, as well as efficient handling of errors and minimization of network utilization.
    

    \item \textbf{FilingDocument}: Each Filing may contain one or more documents, and the FilingDocument record is the base object for this data.  Filing documents are indicated through SGML tags in raw filings, and typically include their own metadata in SGML.  Filing documents are keyed by filing and by sequence number, and are searchable through other metadata such as document type, description, SHA-1 hash, or MIME content type.  
    
    \item \textbf{SearchQuery, SearchQueryTerm, SearchQueryResult}: The SearchQuery objects provide a basic example of extending and customizing the skeleton data model. They also provide out-of-the-box deep text search capability, as many researchers simply want to identify documents that reference certain concepts for subsequent hand-coding or review.
\end{itemize}

\subsection{Clients}
OpenEDGAR provides high-level clients for interacting with the object storage engine and EDGAR itself.  While both are HTTP-based APIs, these high-level clients substantially simplify the development and customization of OpenEDGAR as described below.

\begin{itemize}
    \item \textbf{Object Storage}: OpenEDGAR provides a high-level client for object storage through the \textsc{boto} library.  Boto, designed for use with Amazon S3 or compatible engines such as OpenStack Swift, provides low level client interfaces.  We have implemented additional functionality and a number of common patterns into our client, including, for example, transparent document compression and access to document byte ranges.
    
    \item \textbf{EDGAR}: OpenEDGAR provides a high-level client for accessing EDGAR using the \textsc{requests} library.  Functionality includes retrieving file or directory contents, listing and retrieving EDGAR indexes by type or year, and retrieving company metadata.
\end{itemize}

\subsection{Parsers}
The EDGAR system was designed in the early 1990s, prior to the adoption of XML or other modern data interchange formats.  The technologies in use at the time include fixed-width ``flat'' files and SGML documents, and EDGAR still utilizes both.  OpenEDGAR provides parsers that can handle these files, as described below.

\begin{itemize}
    \item \textbf{Index Parser}: EDGAR index files are gzip-compressed fixed-width ``flat'' files.  The columns for each file are dynamically sized based on the maximum length of each value, and we rely on preprocessing and the \textsc{pandas} package to process these files.  Some files have historically been malformed, omitted columns, or have been compressed incorrectly; OpenEDGAR log and handles exceptions such as these, ensuring that fields such as CIK, Company Name, Date Filed, File Name, and Form Type are present.  The index parser returns these results as a \textsc{pandas} DataFrame object.
    
    \item \textbf{Filing Parser}: EDGAR filing are SGML documents that contain a header and one or more documents.  Headers are identified by either \textsc{SEC-HEADER} or \textsc{IMS-HEADER} tags, depending on their age, and each document is identified by a separate \textsc{DOCUMENT} tag.  Filing headers may include such data as accession number, company name, CIK, and SIC, reporting period, and date; however, data quality varies substantially across time, form type, filer, and filer type.  The filing parser returns a Python dictionary object containing metadata and a list of parsed Filing Documents, as returned by the Filing Document parser described below.
    
    \item \textbf{Filing Document Parser}: As described above, each EDGAR filing contains one or more filing documents identified by an SGML \textsc{DOCUMENT} tag.  Each document may contain its own metadata such as sequence, type, description, and file name, as well as the document contents.  Documents contain SGML tags that typically indicate the content type, such as \textsc{PDF}, \textsc{HTML}, or \textsc{XML}, but data quality varies widely.  Additionally, files such as images or PDFs may be uuencoded, although uuencoding is not standards-based and some documents may be malformed.  OpenEDGAR verifies content type and transparently handles uudecoding content.
\end{itemize}

\subsection{Processes}
The data model, clients, and parsers provide all the necessary pieces to construct research database from EDGAR.  While the process or workflow of each research project may vary, OpenEDGAR also provides ``standard'' processes or workflows to accomplish extremely common tasks.

\begin{itemize}
    \item \textbf{EDGAR}: OpenEDGAR's \textsc{processes.edgar} and \textsc{tasks} modules include methods that:
    \begin{itemize}
        \item initially populate database objects for company metadata
        \item incrementally populate database objects for company metadata
        \item initially download filing index files
        \item incrementally download filing index files
        \item initially download filing files
        \item incrementally download filing files
        \item initially populate database objects from filing index and filings
        \item incrementally update database objects filing index and filings
        \item initially extract text content from filing documents
        \item incrementally extract text content from filing documents
        \item search filing documents for term references
    \end{itemize}
    
    \item \textbf{S3}:OpenEDGAR's \textsc{processes.s3} module include methods that:
    \begin{itemize}
        \item locate and remove rate-limited objects
        \item locate and remove empty objects
        \item locate and remove access denied objects
    \end{itemize}
\end{itemize}

Interested readers are directed to the \textsc{openedgar} at \url{https://github.com/LexPredict/openedgar} on GitHub for more details and source code. 

\section{Introductory OpenEDGAR Examples}
\label{section:examples}

While there are many examples of research relying on EDGAR generally, we provide three example usages of OpenEDGAR, including development guidelines and excerpted source code.  These examples include measuring regulatory references in annual 10-K filings, locating clauses in employment agreements, and training doc2vec word embedding models from press releases.  While these examples are fairly simple, the OpenEDGAR framework combined with other open and closed source tools can support a range of more sophisticated efforts.  

\subsection{Measuring trends in regulatory references}
\label{subsection:example_10k}
Our first example is based on previous academic research conducted by two of this article's authors in \cite{BommaritoII2017}, in which we analyzed the annual 10-K reports of over 34,000 filing companies for 23 years.  We examined these Form 10-K reports for references to U.S. Federal acts and agencies, building a database of over 4.5 million records spanning hundreds of regulations and regulators.  In order to replicate this analysis or carry out similar research, it is critical that scholars have access to 10-K filings and their textual content.

OpenEDGAR makes this task as simple as three lines of code without any additional customization or development.  Once the application has been installed as documented in the Installation documentation, a user can execute the code listed in \ref{code:example_10k}.  This code can be run from a Python shell or Jupyter notebook.

These methods execute in parallel using as many cores or machines as configured in celery.  As of May 2018, when executed on an \textit{m5.large} EC2 instance at AWS with 2 cores and 8GB RAM, the initial data retrieval, database population, and content extraction steps take approximately 24 hours to complete.  As documented in table \ref{tab:example_2018_10k}, there are over 2,000 filings, 200,000 filing documents, and 5GB of compressed data on S3 once these commands finish executing.  By removing the \mintinline{python}{year=2018} argument from lines 2 and 3, the entire history of EDGAR can be retrieved, although this will take much longer to complete and requires substantially more storage.

\begin{table}[ht]
    \centering
    \begin{tabular}{|c|c|}
         \hline
         Metric & Value \\\hline
         Number of Filers & 2,080\\\hline
         Number of Filings & 2,175\\\hline
         Number of Filing Documents & 224,187\\\hline
         Time to Run & 24 hrs\\\hline
         Size of Raw Documents & 3.7GB\\\hline
         Size of Extracted Text & 1.7GB\\\hline
    \end{tabular}
    \caption{Example statistics for 2018 10-K retrieval and parsing with OpenEDGAR}
    \label{tab:example_2018_10k}
\end{table}

Our research in \cite{BommaritoII2017} can be replicated by extending OpenEDGAR's data model and tasks as follows:
\begin{enumerate}
    \item Extend the data model to create objects for Act and Act Aliases.
    \item Populate the Act and Act Alias tables with the desired acts to analyze.
    \item Extend the data model to create objects for Agency and Agency Aliases.
    \item Populate the Agency and Agency Alias tables with desired agencies to analyze.
    \item Extend the data model to create objects for Act References and Agency references, which contain foreign keys to Act or Agency and Filing Document.
    \item Depending on the researcher's tolerance for recall or precision, either:
    \begin{enumerate}
        \item Obtain lower recall with less compute time using the exact term search functionality in OpenEDGAR to locate Act and Agency References.
        \item Obtain higher recall with more compute time using a natural language processing library such as LexNLP or NLTK to locate Act and Agency References (\cite{BirdKleinLoper09, lexnlp2018}).
    \end{enumerate}
\end{enumerate}

\begin{figure}[ht!]
    \centering
    \includegraphics[width=6.9in]{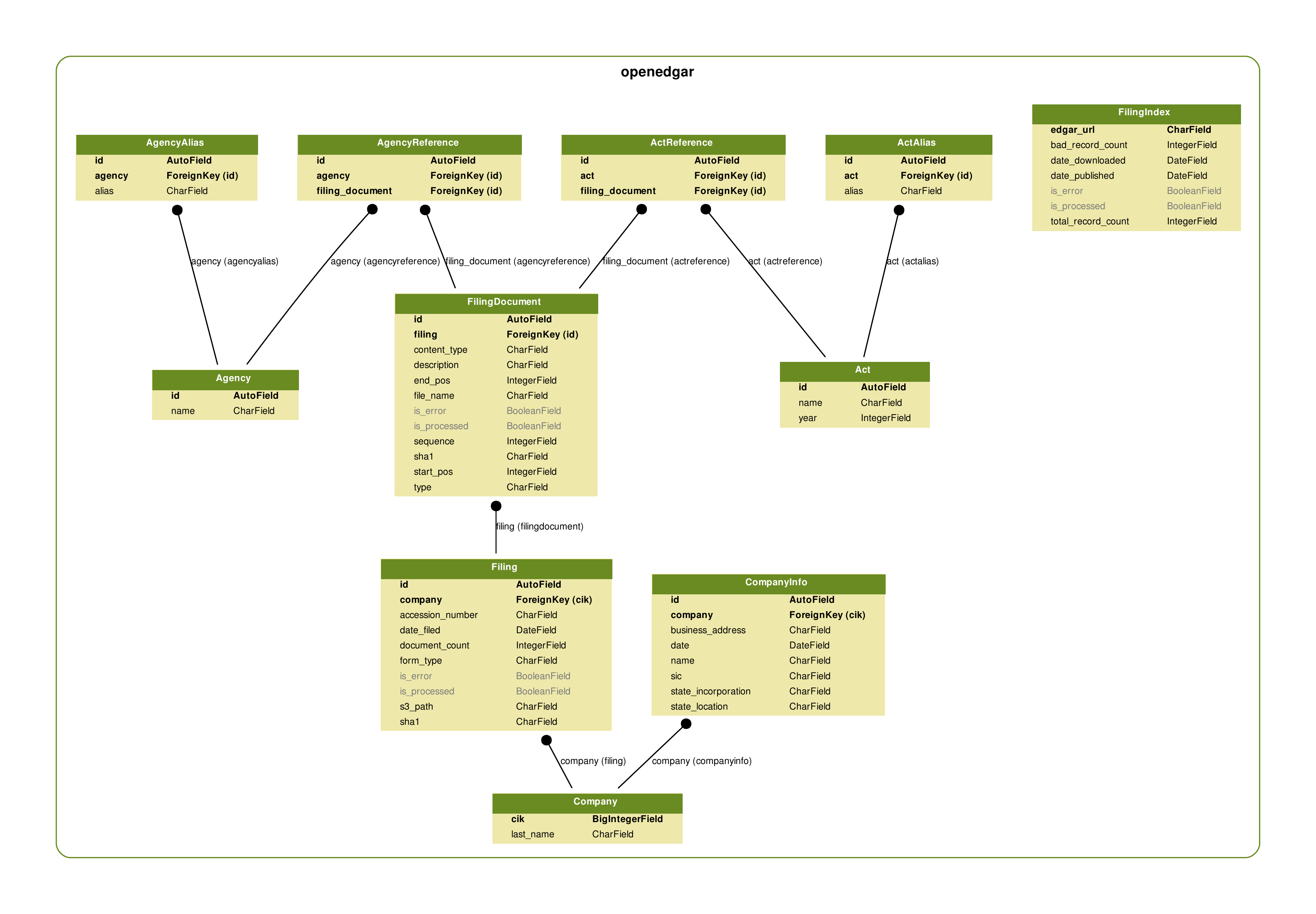}
    \caption{Data model diagram for a 10-K act and agency reference research}
    \label{fig:openedgar_10k_data_model}
\end{figure}

Figure \ref{fig:openedgar_10k_data_model} provides a UML representation of the resulting customized OpenEDGAR data model.  We estimate that the data in \cite{BommaritoII2017} can be approximately replicated with lower recall by adding fewer than 50 lines of code to OpenEDGAR.

\subsection{Identifying non-compete clauses in executive employment agreements}
Scholars in law and finance have often studied agreements drawn from EDGAR, including credit agreements, employment agreements, and merger agreements (\cite{schwab2006empirical, eisenberg2006ex, nini2009creditor, roberts2009renegotiation, sanga2014choice, rauterberg2017contracting}).  In this example, we will demonstrate how a researcher could quickly retrieve a sample of employment agreements and review them for the presence of non-compete and/or non-solicitation clauses.


Agreements and contracts make their way into EDGAR through a variety of form types and conditions, as detailed in regulatory guidance like 17 CFR 229.601.  As detailed in (b)(10)(iii)(A) of 229.601, material contracts include:

\begin{quote}
(A) Any management contract or any compensatory plan, contract or arrangement, including but not limited to plans relating to options, warrants or rights, pension, retirement or deferred compensation or bonus, incentive or profit sharing (or if not set forth in any formal document, a written description thereof) in which any director or any of the named executive officers of the registrant, as defined by Item 402(a)(3) (\S 229.402(a)(3)), participates shall be deemed material and shall be filed; and any other management contract or any other compensatory plan, contract, or arrangement in which any other executive officer of the registrant participates shall be filed \textit{unless immaterial in amount or significance} (emphasis added).
\end{quote}

Thus, most filers are required to publish most executive employment agreements, both periodically and as driven by triggering events.  We can search for these agreements in EDGAR in a number of ways, but in our experience, the vast majority are found in 8-K, 10-Q, or 10-K filings.  For simplicity, the code listed in \ref{code:example_non_solicit} shows results from the 2018 10-K filings retrieved in the first case study above; a more complete investigation of this question would require analysis of other filing types and longer period.  The code, however is unchanged between those forms of analysis, and can be executed on Python shell or in a Jupyter notebook.

In plain English, the code listed in \ref{code:example_non_solicit} first uses the Django ORM to query \mintinline{python}{FilingDocument} objects for ones whose description includes ``employment agreement'' (case-insensitive).  Many, but not all, agreements contain accurate metadata parsed from the SGML headers described above, and for this example, we rely on this metadata alone.  From a recall perspective, however, it should be noted that many agreements will not be identified without examining the contents of the filing document directly.  Researchers who desire to obtain all EDGAR-filed agreements should implement or use document classifiers like those in \cite{lexnlp2018}.

Once the ORM returns a result set of relevant filing documents, we next retrieve the text contents of each document.  The text contents is produced at document ingestion by the code in example \ref{subsection:example_10k} above and stored in object storage, where we now retrieve it by SHA-1 hash.  As some documents may contain binary or non-English characters, the OpenEDGAR storage client returns byte streams; for this example, however, we simply decode as UTF-8.

Finally, for each agreement, we use LexNLP(\cite{lexnlp2018}) to extract the stems for each sentence, counting the number of ``solicit'' stem occurrences per agreement.  This count is tracked with basic document metadata and then converted to a \textsc{pandas} data frame to enable subsequent analysis.  While a range of more sophisticated extraction and clause classification protocols can be developed leveraging \cite{lexnlp2018} and other open and closed source tools, we provide this simple example as an illustrative starting point.  

\begin{figure}[ht!]
    \centering
    \includegraphics[width=3.5in]{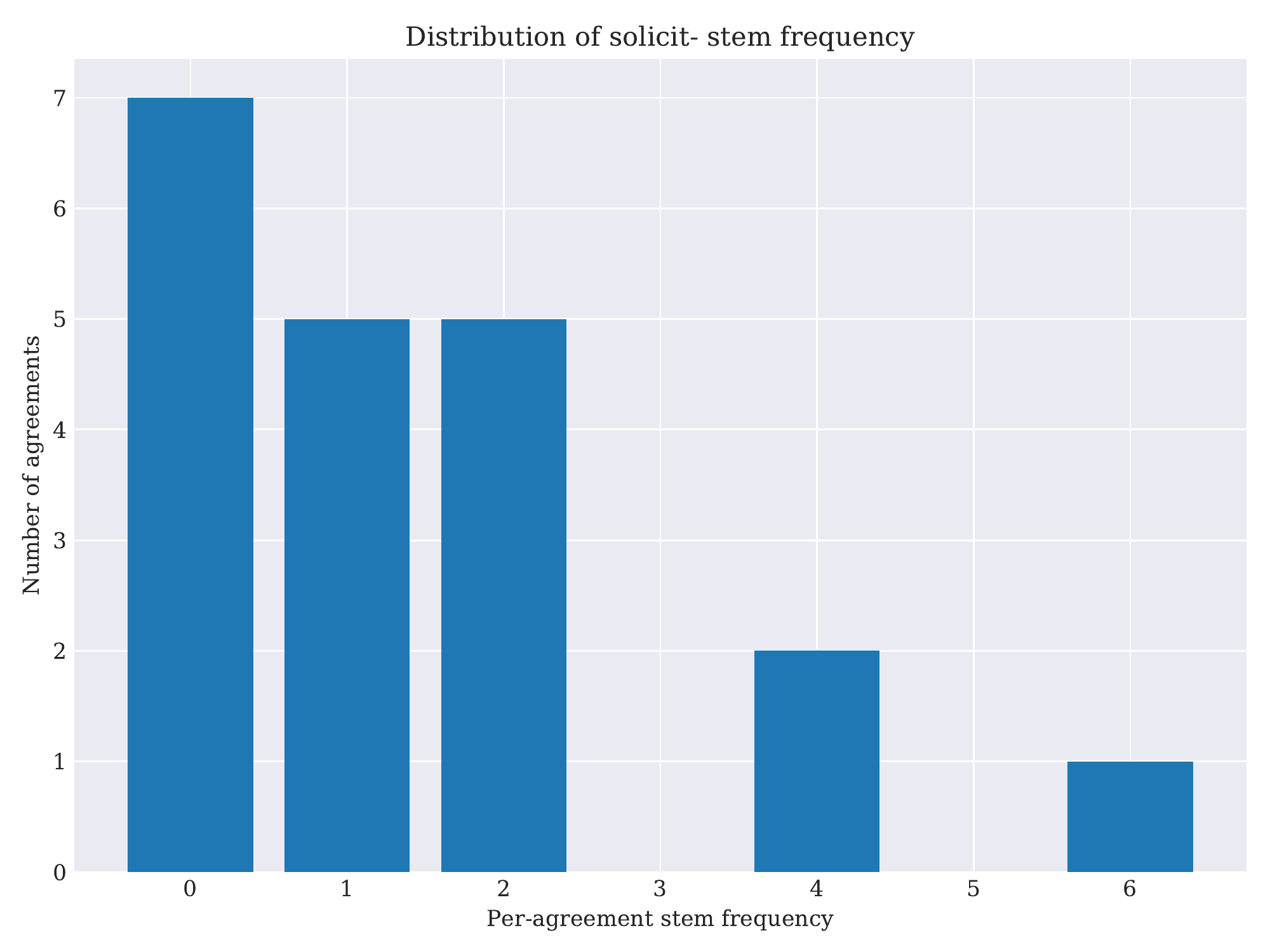}
    \caption{Distribution of solicit- stem frequency per agreement}
    \label{fig:employment_example_solicit_histogram}
\end{figure}

We can then visualize data in the \mintinline{python}{solicit_df} data frame either within the Jupyter notebook or by exporting figures.  Figure \ref{fig:employment_example_solicit_histogram} shows, for example, a histogram of the stem frequency by agreement for the ``solicit-'' concept.  In this limited example, we can see that 7 of 20 (35\%) of agreements do not contain any mention of ``solicit,'' ``solicitation,'' or other solicit- word.  However, the other 13 agreements have at least one occurrence, and 8 agreements have at least two occurrences.

\subsection{Building word embedding models from press releases}
Researchers in both academic and commercial contexts frequently require corpora for the development of natural language and machine learning models.  One common task is the development of word frequency or word embedding models, which allow researchers to encode text or entire documents as vectors in a feature space of words or concepts (\cite{mikolov2013efficient, mikolov2013distributed, ma2015using, liu2016transferring, nay2017predicting}).  For example, in the prior example related to employment agreements, some agreements use the phrase ``divert'' or ``diversion'' instead of ``solicit'' or ``solicitation.''  Such synonyms or related concepts are missed when matching tokens and even when stemming or lemmatizing.  Word embedding models trained with sufficient data, however, often can capture synonyms and related concepts without the need for manual enumeration by researchers.

In the example code listed in \ref{code:example_word2vec}, we demonstrate how researchers can quickly retrieve a sample of 1,000 press releases to build a \textsc{word2vec} word embedding model.  Press releases are generally contained in 8-K filings, and the example code in listing \ref{code:example_10k} can be adapted to retrieve these filings by simply adding \mintinline{python}{"8-K"} to the \mintinline{python}{form_type_list} argument of \mintinline{python}{process_all_filing_index}. Over the window in question, there are over 1.5 million 8-K filings with approximately 6 million filing documents, so researchers should plan accordingly if they intend to work with 8-K data.

Once trained, this model can be be used to produce vector representations of new text or queried to produce synonyms or related concepts.  Table \ref{tab:example_word2vec_revenue} shows the top three stems related to the word ``revenue'' for a sample from our production database.

\begin{table}[ht]
    \centering
    \begin{tabular}{|c|c|c|}
        \hline
        \textbf{Rank} & \textbf{Stem} & \textbf{Similarity} \\\hline
        1 & segment &  0.589056 \\\hline
        2 & sale &  0.575444 \\\hline
        3 & profit &  0.563288 \\\hline
    \end{tabular}
    \caption{Sample word2vec results for ``revenue''; see listing \ref{code:example_word2vec}}
    \label{tab:example_word2vec_revenue}
\end{table}

For researchers who need pretrained word embedding models for legal or regulatory text, the LexNLP package (\cite{lexnlp2018}) includes a number of pre-trained word2vec and doc2vec models.  In forthcoming research, we will also be releasing a large-scale data set and ``genomic'' model of contracts and their clauses, including contract data models and associated classifiers.

\section{Acknowledgements}
\label{section:acknowledgments}
We would like to acknowledge the support of our company, LexPredict, and the developers and analysts who have helped in the design, development, maintenance, and testing of this software.  We would also like to acknowledge the contribution of the teams behind the Django, celery, Apache Tika, and Python itself; without the ecosystem created by their work, this software would not exist.
\\
\\



\bibliographystyle{model1-num-names}
\bibliography{openedgar.bib}


\appendix
\onecolumn
\section{Source Code}
\subsection{Building a 2018 10-K database}
\label{code:example_10k}
\begin{minted}[breaklines,frame=lines,fontsize=\footnotesize]{python}
from openedgar.processes.edgar import download_filing_index_data, process_all_filing_index
download_filing_index_data(year=2018)
process_all_filing_index(year=2018, form_type_list=["10-K"])
\end{minted}

\subsection{Examining non-solicitation clauses in employment agreements}
\label{code:example_non_solicit}
\begin{minted}[breaklines,frame=lines,fontsize=\footnotesize]{python}
import pandas
from lexnlp.nlp.en.segments.sentences import get_sentence_list
from lexnlp.nlp.en.tokens import get_stem_list
from openedgar.models import FilingDocument
from openedgar.clients.s3 import get_buffer

# Use Django ORM to retrieve sample of agreements
sample_size = 20
search_string = "employment agreement"
agreement_fd_list = FilingDocument.objects\
    .filter(description__icontains=search_string)[0:sample_size]

# Retrieve text contents of each agreement and track data
solicit_data = []
for agreement_fd in agreement_fd_list:
    agreement_path = "documents/text/{0}".format(agreement_fd.sha1)
    agreement_contents = get_buffer(agreement_path).decode('utf-8')
    
    # Use LexNLP to loop through sentences and match on stem, not token
    solicit_count = 0
    for sentence in get_sentence_list(agreement_contents):
        stems = get_stem_list(sentence, lowercase=True)
        solicit_count += stems.count("solicit")
    
    # Append record for agreement
    solicit_data.append({"sha1": agreement_fd.sha1,
                         "description": agreement_fd.description,
                         "solicit_count": solicit_count})
        
# Use pandas to show histogram
solicit_df = pandas.DataFrame(solicit_data)
\end{minted}

\subsection{Training and using a word2vec model from press releases}
\label{code:example_word2vec}
\begin{minted}[breaklines,frame=lines,fontsize=\footnotesize]{python}
from lexnlp.nlp.en.segments.sentences import get_sentence_list
from lexnlp.nlp.en.tokens import get_stem_list, DEFAULT_STEMMER
from openedgar.models import FilingDocument
from openedgar.clients.s3 import get_buffer

# Use Django ORM to retrieve sample of agreements
sample_size = 1000
search_string = "press release"
release_fd_list = FilingDocument.objects\
    .filter(description__icontains=search_string)[0:sample_size]

# Retrieve text contents of each agreement and track data
sentence_list = []
for release_fd in release_fd_list:
    release_path = "documents/text/{0}".format(release_fd.sha1)
    release_contents = get_buffer(release_path).decode('utf-8')
    
    # Use LexNLP to loop through sentences and build stopworded sentence stem list
    for sentence in get_sentence_list(release_contents):
        stems = get_stem_list(sentence, lowercase=True, stopword=True)
        sentence_list.append(stems)
\end{minted}

\begin{minted}[breaklines,frame=lines,fontsize=\footnotesize]{python}
import gensim.models.word2vec
word2vec_model = gensim.models.word2vec.Word2Vec(sentence_list)
word2vec_model.wv.most_similar(positive=[DEFAULT_STEMMER.stem("revenue")], topn=3)
\end{minted}
\end{document}